%% file: main.tex
\documentclass[10pt,twocolumn,letterpaper]{article}

\usepackage[pagenumbers]{cvpr} 

\input{preamble}

\definecolor{cvprblue}{rgb}{0.21,0.49,0.74}
\usepackage[pagebackref,breaklinks,colorlinks,citecolor=cvprblue]{hyperref}
\usepackage{booktabs}
\usepackage{colortbl}
\usepackage{amssymb}
\usepackage{booktabs}

\def\paperID{*****} 
\def\confName{CVPR}
\def\confYear{2024}

\title{DEYO: DETR with YOLO for End-to-End Object Detection}

\author{Haodong Ouyang\\
Southwest Minzu University\\
Chengdu, China\\
{\tt\small ouyanghaodong@stu.swun.edu.cn}}

\begin{document}
\maketitle
\input{sec/0_abstract}    
\input{sec/1_intro}
\input{sec/2_rel}
\input{sec/3_deyo}
\input{sec/4_exp}

\input{sec/5_col}

{
    \small
    \bibliographystyle{ieeenat_fullname}
    \bibliography{main}
}


\end{document}

%% file: preamble.tex
\usepackage[dvipsnames]{xcolor}


%% file: sec/0_abstract.tex
\begin{abstract}
The training paradigm of DETRs is heavily contingent upon pre-training their backbone on the ImageNet dataset. However, the limited supervisory signals provided by the image classification task and one-to-one matching strategy result in an inadequately pre-trained neck for DETRs. Additionally, the instability of matching in the early stages of training engenders inconsistencies in the optimization objectives of DETRs. To address these issues, we have devised an innovative training methodology termed step-by-step training. Specifically, in the first stage of training, we employ a classic detector, pre-trained with a one-to-many matching strategy, to initialize the backbone and neck of the end-to-end detector. In the second stage of training, we froze the backbone and neck of the end-to-end detector, necessitating the training of the decoder from scratch. Through the application of step-by-step training, we have introduced the first real-time end-to-end object detection model that utilizes a purely convolutional structure encoder, DETR with YOLO (DEYO). Without reliance on any supplementary training data, DEYO surpasses all existing real-time object detectors in both speed and accuracy. Moreover, the comprehensive DEYO series can complete its second-phase training on the COCO dataset using a single 8GB RTX 4060 GPU, significantly reducing the training expenditure. Source code and pre-trained models are available at \url{https://github.com/ouyanghaodong/DEYO}.
\end{abstract}

%% file: sec/1_intro.tex
\section{Introduction}
\label{sec:intro}

\begin{figure}[t]
\vspace{3mm}
  \centering
  \includegraphics[width=0.88\linewidth]{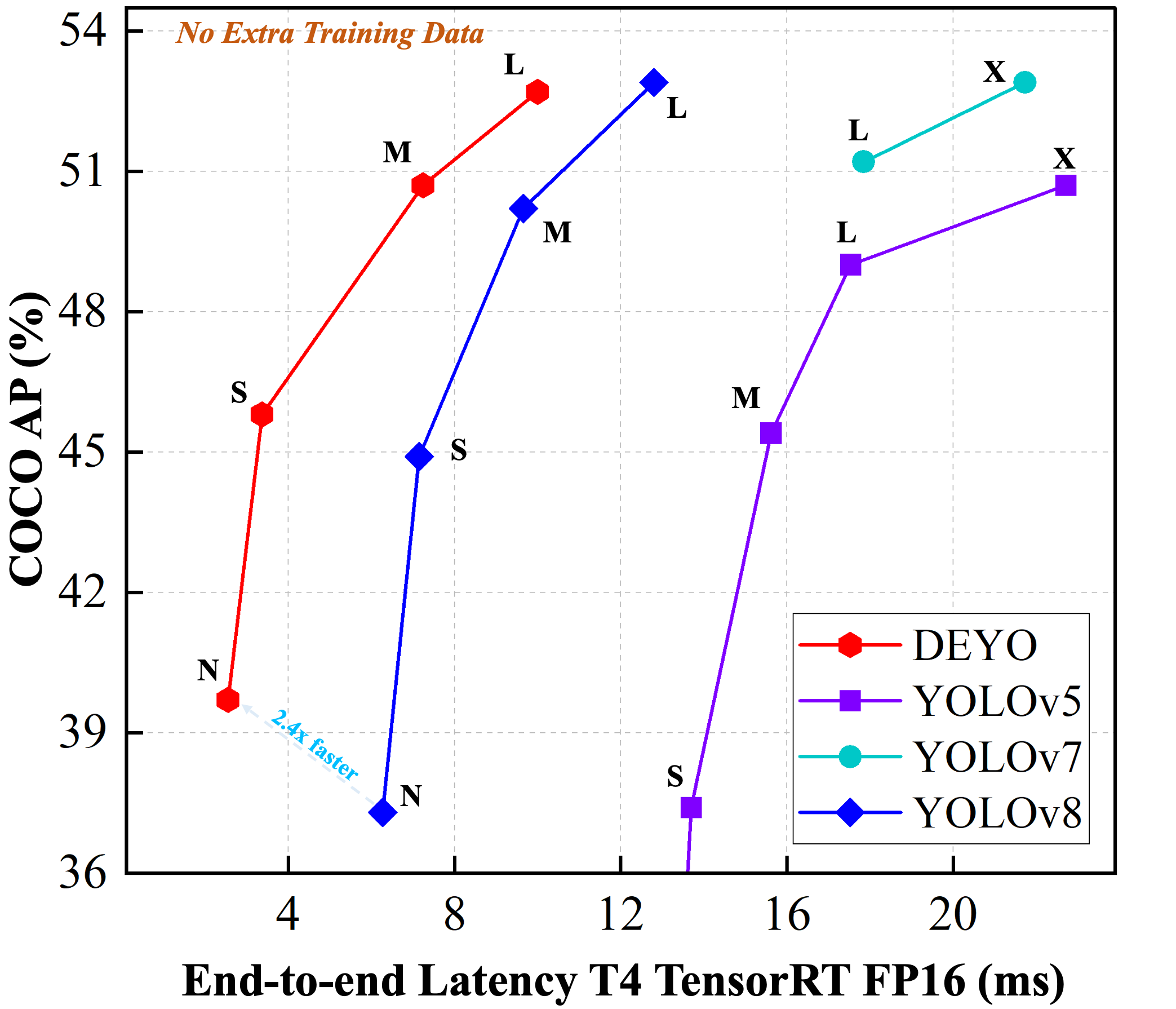}
   \caption{DEYO has surpassed other real-time object detectors in speed and accuracy; all detectors were exclusively trained on the COCO dataset without any additional datasets.}
   \label{fig:1}
\vspace{-3mm}
\end{figure}

\hspace{1pc}Object detection is a fundamental task within the field of computer vision, tasked with the precise localization and identification of various object categories within images or videos. This technology is a cornerstone for many computer vision applications, including autonomous driving, video surveillance, facial recognition, and object tracking. In recent years, advancements in deep learning, particularly methods based on Convolutional Neural Networks (CNNs) \cite{64}, have led to groundbreaking progress in object detection tasks, establishing themselves as the predominant technology in this domain. 

DEtection TRansformer (DETR) \cite{1} introduces an end-to-end approach for object detection, comprising a CNN backbone, transformer encoder, and transformer decoder. DETR employs a Hungarian loss to predict a one-to-one set of objects, thereby eliminating reliance on the manually tuned component of Non-Maximum Suppression (NMS), which significantly streamlines the object detection pipeline through end-to-end optimization.

Although end-to-end object detectors based on Transformers (DETRs) have achieved notable success in terms of performance, these detectors typically rely on pre-training their backbone networks on the ImageNet dataset. Should a new backbone be selected, it necessitates pre-training on ImageNet before training the DETRs or utilizing an existing pre-trained backbone. Such dependency limits the flexibility in designing the backbone and escalates development costs, and when the task dataset significantly diverges from ImageNet, this pre-training strategy may result in suboptimal fine-tuning outcomes for DETRs on specific datasets.

Furthermore, since DETRs employ the Hungarian matching algorithm for direct one-to-one set prediction of objects, and the complexity of their decoder is quadratic in relation to the length of the input sequence, the number of queries receiving direct supervision signals during training is substantially less than that in classic object detectors using a one-to-many matching strategy. Coupled with the inherent limitations of image classification tasks, this results in the neck of DETRs not being sufficiently pre-trained. Additionally, during the early stages of DETR \cite{1} training, the same query often matches with different objects at different times within the same image, leading to an optimization process that is both ambiguous and unstable, thereby undermining the pre-trained backbone.

To address the aforementioned challenges, we introduce an innovative training paradigm termed "step-by-step training." This approach commences with a pre-training phase on a custom dataset, utilizing a classic detector for the task of object detection, thereby circumventing the need for additional datasets. Subsequently, in the second phase of training, the backbone and neck of the classically trained detector refined through one-to-many matching during the initial phase are employed to initialize an end-to-end detector. During this phase, the backbone and neck components of the end-to-end detector are frozen, allowing for the exclusive retraining of the decoder from scratch. The step-by-step training approach yields a notable enhancement in performance compared to the conventional training methodology of DETRs. Concurrently, this step-by-step training substantially reduces the training costs for the detector: the first phase of training can be completed with just 16GB of VRAM, while the second phase requires a mere 8GB of VRAM.

Leveraging a step-by-step training approach, we introduce the first real-time end-to-end object detector employing a purely convolutional architecture as the encoder, named DETR \cite{1} with YOLO \cite{4, 5, 6} (DEYO). Specifically, we commence by training a robust YOLO object detection model on custom datasets to initialize the backbone and neck of DEYO. Subsequently, we combine the pretrained neck with a straightforward feature projection to construct DEYO's lightweight decoder. Owing to the high-quality pretraining provided to DEYO's backbone and neck in the initial phase, DEYO surpasses contemporary state-of-the-art real-time object detectors in terms of speed and accuracy. 

DEYO-tiny achieves 37.6\% AP on COCO \cite{12} \texttt{val2017} and operates at 497 FPS on the NVIDIA Tesla T4 GPU, while DEYO-X attains 53.7\% AP and 65 FPS. Furthermore, by discarding the reliance on NMS, DEYO demonstrates a notable performance enhancement over YOLOv8 \cite{62} on the CrowdHuman \cite{63} dataset. Without additional training data, DEYO outperforms all comparable real-time detectors in speed and precision, establishing a new state-of-the-art for real-time object detection.

The main contributions of this paper are summarized as follows:

\begin{enumerate}
    \item We propose the first training method that does not require additional datasets to train DETRs: step-by-step training. Compared with conventional training methods for DETRs, step-by-step training can provide high-quality pre-training for the detector's neck and fundamentally solve the damage to the backbone due to unstable binary matching in the early stage of training, thereby significantly improving the performance of the detector.
    
    \item Using step-by-step training, we develop the first real-time end-to-end object detector DEYO using a purely convolutional structure as the encoder, which surpasses the current state-of-the-art real-time detectors in both speed and accuracy, and no post-processing is required, so its inference speed is lag-free and stable.
    
    \item We conduct a series of ablation studies to analyze the effectiveness of our proposed method and the model's different components.
\end{enumerate}

%% file: sec/2_rel.tex
\section{Related Work}
\label{sec:rel}

\begin{figure*}[t]
\begin{center}
\includegraphics[width=0.92\linewidth]{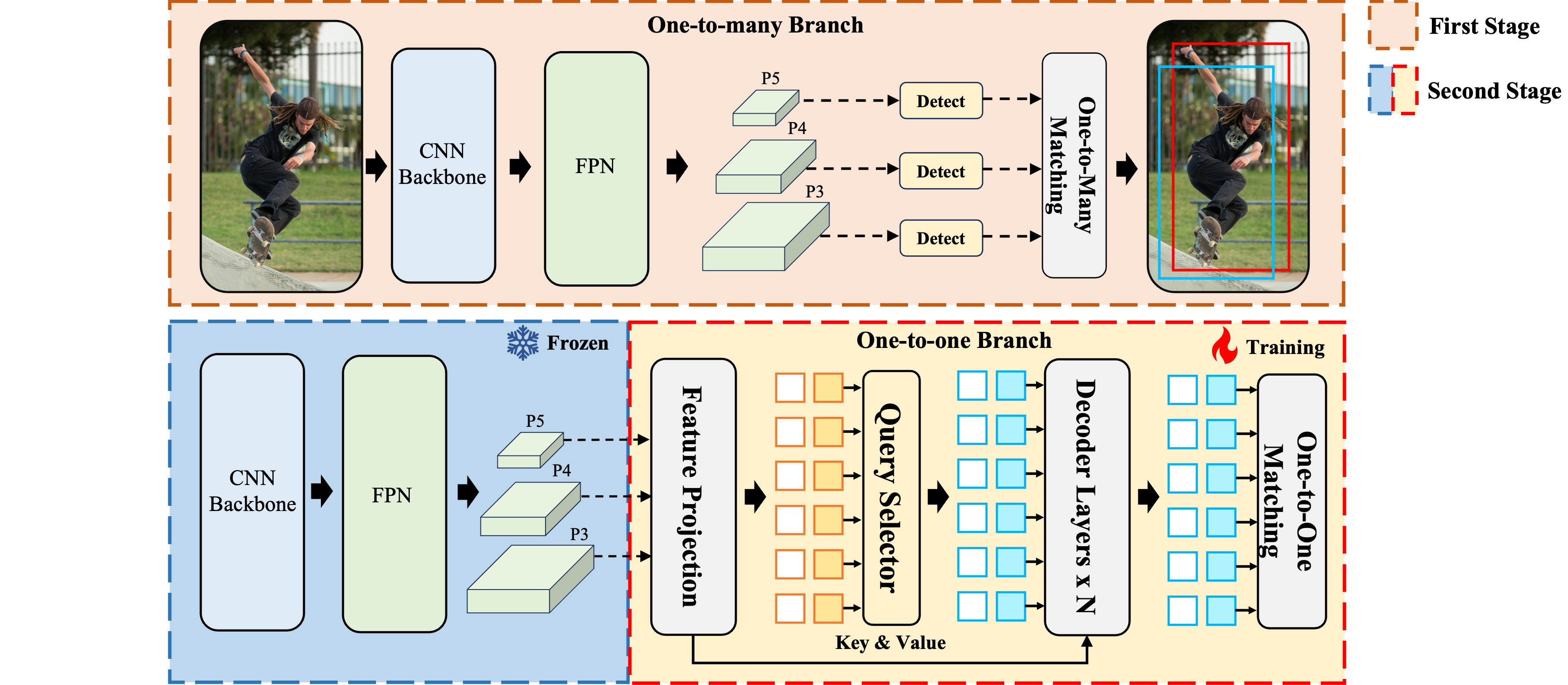}
   \caption{We eliminated the encoder usage and instead employed the multi-scale features \{P3, P4, P5\} provided by the neck. Following feature projection, these features were utilized as input for the encoder while simultaneously generating candidate bounding boxes and filtering them through the query selector. Subsequently, this information was passed into a decoder with an auxiliary prediction head, enabling iterative optimization for generating bounding boxes and scores.}
   \label{fig:2}
\vspace{-4mm}
\end{center}
\end{figure*}

\subsection{DEtection TRansformers (DETR)}
\hspace{1pc}Carion et al. proposed an end-to-end object detector based on transformers, named DETR (DEtection TRans- former) \cite{1}, which has attracted significant attention from researchers due to its end-to-end nature in object detection. Specifically, DETR eliminates the anchor and NMS components in traditional detection pipelines and adopts a bipartite graph matching label assignment method to directly predict one-to-one sets of objects. This strategy dramatically simplifies the object detection process and alleviates the performance bottleneck caused by NMS. However, DETR suffers from slow convergence speed and query ambiguity issues. To address these problems, several variants of DETR have been proposed, such as Deformable-DETR \cite{7}, Conditional-DETR \cite{22}, Anchor-DETR \cite{46}, DAB-DETR \cite{8}, DN-DETR \cite{13}, and DINO \cite{14}. Deformable-DETR enhances the efficiency of attention mechanisms and accelerates training convergence by utilizing multi-scale features. Conditional-DETR and Anchor-DETR reduce the optimization difficulty of queries. DAB-DETR introduces 4D reference points and optimizes predicted boxes layer by layer. DN-DETR speeds up training convergence by introducing query denoising. DINO improves upon previous work and achieves state-of-the-art results. However, the aforementioned improvements do not address the issue of high computational cost in DETR. RT-DETR \cite{57} designs an efficient hybrid encoder to replace the original transformer encoder, reducing unnecessary computational redundancy in the DETR encoder and proposing the first end-to-end object detector.

\subsection{You Only Look Once (YOLO)}
\hspace{1pc}Over the years, the YOLO \cite{4,5,6} series has been one of the best single-stage real-time object detector categories. YOLO transforms the object detection task into a regression problem, predicting the positions and categories of multiple objects in a single forward pass, achieving high-speed object detection. After years of development, YOLO has developed into a series of fast models with good performance. Anchor-based YOLO methods include YOLOv4 \cite{25}, YOLOv5 \cite{61}, and YOLOv7 \cite{26}, while anchor-free methods are YOLOX \cite{27}, YOLOv6 \cite{29}, and YOLOv8 \cite{62}. Considering the performance of these detectors, anchor-free methods perform as well as anchor-based methods, and anchor boxes are no longer the main factor limiting the development of YOLO. However, all YOLO variants generate many redundant bounding boxes, which NMS must filter out during the prediction stage, which significantly impacts the detector's accuracy and speed and conflicts with the design theory of real-time object detectors. 

%% file: sec/3_deyo.tex
\section{DEYO}
\label{sec:deyo}

\subsection{Model Overview}
\hspace{1pc}Fig~\ref{fig:2} illustrates the comprehensive architecture of our proposed DEYO. DEYO employs YOLOv8 \cite{62} as its one-to-many branch, wherein YOLOv8 comprises a backbone, a Feature Pyramid Network (FPN) \cite{34}, and a Path Aggregation Network (PAN) \cite{35} that together form the neck structure, in addition to a head capable of producing predictions at three different scales. Conversely, DEYO's one-to-one branch utilizes a lightweight, purely convolutional encoder and a Transformer-based decoder. Moreover, we have also incorporated a CDN component identical to that used in DINO \cite{14} to enhance the model's precision.

\subsection{One-to-many Branch}
\hspace{1pc}The YOLO \cite{4,5,6} model's generalization capabilities and practicality have been extensively validated and widely acknowledged within the field of computer vision. Even without the aid of additional datasets, YOLO demonstrates exceptional performance in processing complex scenes, executing multi-object detection, and adapting to real-time applications. Leveraging these benefits, we selected YOLO as the one-to-many branch for our DEYO model, providing DEYO with a high-quality, pre-trained backbone and neck structure. This branch features three multi-scale output layers capable of generating up to 8,400 candidate regions. Unlike the one-to-one label assignment strategy adopted by the DETR model, YOLO benefits from a one-to-many label assignment strategy during its training process, which, due to a higher quantity of positive samples, offers more comprehensive supervision of the network in its initial training stages. These candidate regions are tasked with more than mere classification; they confront the more complex challenge of object detection. This further cultivates a robust neck structure, supplying the decoder with rich multi-scale information, thereby significantly enhancing the overall performance of the model.

\begin{figure}[t]
  \centering
  \includegraphics[width=0.99\linewidth]{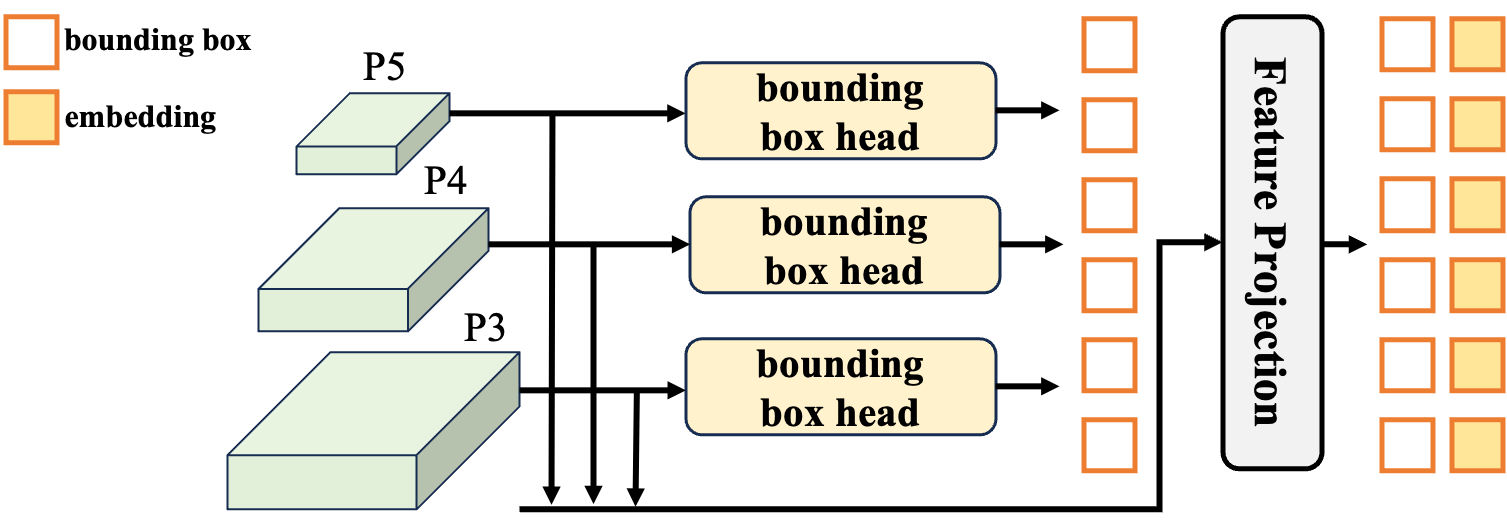}
   \caption{We decouple the generation of bounding boxes from embeddings, allowing for more effective compression of multiscale information emanating from the neck through enhanced feature projection.}
   \label{fig:3}
\end{figure}

\subsection{Efficient Encoder}
\hspace{1pc}Contrary to DETR \cite{1}, which employs a transformer as its encoder, DEYO harnesses the purely convolutional architecture of YOLO's Neck, which is pre-trained in the initial phase to encode multi-scale features. These encoded features are then fed into a feature projection module to align them with the hidden dimensions. Owing to the neck's robust multi-scale feature extraction capabilities, acquired through efficient pretraining at the outset, the encoder can supply the decoder with high-quality keys, values, and proposed bounding boxes. Compared to DETR's randomly initialized multi-scale layers and transformer encoder, DEYO's purely convolutional structure achieves remarkable speed. The process can be summarized as follows:
\begin{align}
\label{eq:eq1} \nonumber
& \textcolor{black}{S_1 = Proj (P_3, P_4, P_5)}\\
& \textcolor{black}{S_2 = Concat(S_1)}\\
\label{eq:eq1} \nonumber
& \textcolor{black}{Q = K = V = S_2}
\end{align}

\subsection{Query Generation}
\hspace{1pc}As illustrated in Fig~\ref{fig:3}, DEYO's query generation method diverges from DETR's conventional two-stage strategy. Specifically, DEYO employs a decoupled generation method for bounding boxes and embeddings, allowing for more efficient compression of multi-scale information from the neck by the feature projection. Concurrently, DEYO inherits a one-to-many branch pre-trained bounding box head, transitioning the learning strategy from dense to sparse rather than training from scratch.

\subsection{One-to-one Branch}
\hspace{1pc}As illustrated, DEYO's one-to-one branch adopts an architecture akin to that of DINO, harnessing the Transformer's self-attention mechanism to capture inter-query relationships, thereby establishing score differentials that suppress redundant bounding boxes. Within each layer of the transformer decoder, the queries are progressively refined, culminating in predictions that correspond on a one-to-one basis with objects. This design significantly streamlines the object detection process within DEYO and eliminates the dependency on Non-Maximum Suppression (NMS), ensuring a consistent inference speed. During the second stage of DEYO's training, we freeze the backbone and neck of DEYO to fundamentally circumvent the instability of bipartite matching during the initial stages of training, which could otherwise detrimentally affect the pretrained backbone. Benefiting from the high-quality initialization provided by the first phase, DEYO achieves rapid convergence and exceptional performance, even when supervising only a few hundred queries in the one-to-one branch and training from scratch.

\begin{figure}[t]
  \centering
  \includegraphics[width=0.78\linewidth]{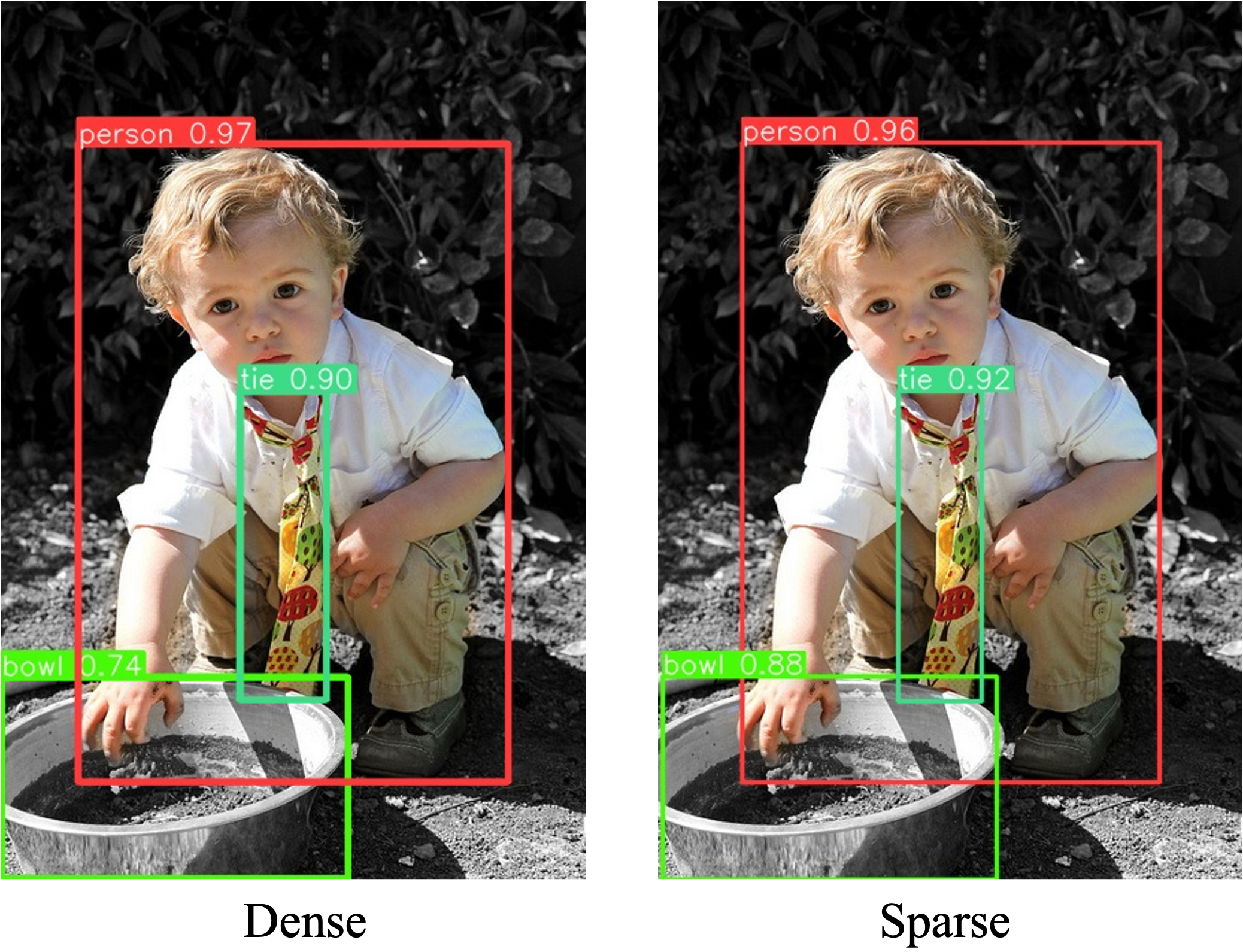}
   \caption{DEYO inherits a one-to-many branch pre-trained bounding box head, transitioning the learning strategy of the bounding box head from dense to sparse rather than training from scratch.}
   \label{fig:4}
\end{figure}

%% file: sec/4_exp.tex
\section{Experiment}
\label{sec:exp}

\begin{table*}[t]
\begin{center}
\resizebox{1.0\textwidth}{!}{%
\begin{tabular}{lccccccccccc}  
\toprule
\bf{Model} & \bf{Backbone} & \bf{Epochs} & \bf{\#Params (M)} & \bf{GFLOPs} & \bf{FPS$_{\text{bs}=1}$} & \bf{AP} & \bf{AP$_{50}$} & \bf{AP$_{75}$} & \bf{AP$_S$} & \bf{AP$_M$} & \bf{AP$_L$} \\
\midrule
\rowcolor{gray!20}
\multicolumn{12}{l}{\textit{Real-time Detectors}} \\
YOLOv5-N \cite{61} & -- & -- & 2 & 5 & 79 & 28.0 & 46.2 & 29.2 & 14.1 & 32.2 & 36.7 \\
YOLOv5-S \cite{61} & -- & -- & 7 & 17 & 76 & 37.4 & 57.2 & 40.2 & 21.1 & 42.3 & 49.0 \\
YOLOv5-M \cite{61} & -- & -- & 21 & 49 & 67 & 45.4 & 64.4 & 48.9 & 27.8 & 50.4 & 58.1 \\
YOLOv5-L \cite{61} & -- & -- & 47 & 109 & 59 & 49.0 & 67.6 & 53.1 & 31.8 & 54.4 & 62.3 \\
YOLOv5-X \cite{61} & -- & -- & 87 & 206 & 44 & 50.7 & 68.9 & 54.6 & 33.8 & 55.7 & 65.0 \\
YOLOv8-N \cite{62} & -- & -- & 3 & 9 & 163 & 37.3 & 52.5 & 40.5 & 18.6 & 41.0 & 53.5 \\
YOLOv8-S \cite{62} & -- & -- & 11 & 29 & 143 & 44.9 & 61.8 & 48.6 & 25.7 & 49.9 & 61.0 \\
YOLOv8-M \cite{62} & -- & -- & 26 & 79 & 106 & 50.2 & 67.2 & 54.6 & 32.0 & 55.8 & 66.4 \\
YOLOv8-L \cite{62} & -- & -- & 44 & 165 & 82 & 52.9 & 69.8 & 57.5 & 35.3 & 58.3 & 69.8 \\
YOLOv8-X \cite{62} & -- & -- & 68 & 258 & 58 & 53.9 & 71.0 & 58.7 & 35.7 & 59.3 & 70.7 \\
\midrule
\rowcolor{gray!20}
\multicolumn{12}{l}{\textit{End-to-end Object Detectors}} \\
DETR \cite{1} & R50 & 500 & 41 & 187 & -- & 43.3 & 63.1 & 45.9 & 22.5 & 47.3 & 61.1 \\
Anchor-DETR \cite{46} & R50 & 50 & 39 & 172 & -- & 44.2 & 64.7 & 47.7 & 23.7 & 49.5 & 62.3 \\
Conditional-DETR \cite{22} & R50 & 108 & 44 & 195 & -- & 45.1 & 65.4 & 48.5 & 25.3 & 49.0 & 62.2 \\
Efficient-DETR \cite{47} & R50 & 36 & 35 & 210 & -- & 45.1 & 63.1 & 49.1 & 28.3 & 48.4 & 59.0 \\
SMCA-DETR \cite{38} & R50 & 108 & 40 & 152 & -- & 45.6 & 65.5 & 49.1 & 25.9 & 49.3 & 62.6 \\
Deformable-DETR \cite{7} & R50 & 50 & 40 & 173 & -- & 46.2 & 65.2 & 50.0 & 28.8 & 49.2 & 61.7 \\
DAB-Deformable-DETR \cite{8} & R50 & 50 & 48 & 195 & -- & 46.9 & 66.0 & 50.8 & 30.1 & 50.4 & 62.5 \\
DN-Deformable-DETR \cite{13} & R50 & 50 & 48 & 195 & -- & 49.5 & 67.6 & 53.8 & 31.3 & 52.6 & 65.4 \\
DINO \cite{14} & R50 & 36 & 47 & 279 & 5 & 50.9 & 69.0 & 55.3 & 34.6 & 54.1 & 64.6 \\
\midrule
\rowcolor{gray!20}
\multicolumn{12}{l}{\textit{Real-time End-to-end Object Detectors}} \\
RT-DETR-R18 \cite{58} & R18 & 72 & 20 & 60 & 240 & 46.5 & 63.8 & -- & -- & -- & -- \\
RT-DETR-R34 \cite{58} & R34 & 72 & 31 & 92 & 172 & 48.9 & 66.8 & -- & -- & -- & -- \\
RT-DETR-R50 \cite{58} & R50 & 72 & 36 & 100 & 120 & 53.1 & 71.3 & 57.7 & 34.8 & 58.0 & 70.0 \\
RT-DETR-R101 \cite{58} & R101 & 72 & 42 & 136 & 78 & 54.3 & 72.7 & 58.6 & 36.0 & 58.8 & 72.1 \\
RT-DETR-L \cite{58} & HGNetv2 & 72 & 32 & 110 & 126 & 53.0 & 71.6 & 57.3 & 34.6 & 57.3 & 71.2 \\
RT-DETR-X \cite{58} & HGNetv2 & 72 & 67 & 234 & 80 & 54.8 & 73.1 & 59.4 & 35.7 & 59.6 & 72.9 \\
\rowcolor{red!10}
\multicolumn{12}{l}{\textit{No Extra Training Data}} \\
DEYO-tiny & -- & 96 & 4 & 8 & 497 & 37.6 & 52.8 & 40.6 & 17.9 & 41.3 & 54.2 \\
DEYO-N & -- & 96 & 6 & 10 & 396 & 39.7 & 55.6 & 42.7 & 20.5 & 43.1 & 56.4 \\
DEYO-S & -- & 96 & 14 & 26 & 299 & 45.8 & 62.9 & 49.3 & 26.8 & 49.9 & 62.5 \\
DEYO-M & -- & 96 & 33 & 78 & 140 & 50.7 & 68.4 & 55.0 & 32.8 & 55.5 & 67.2 \\
DEYO-L & -- & 96 & 51 & 155 & 100 & 52.7 & 70.2 & 57.0 & 36.0 & 57.3 & 69.4 \\
DEYO-X & -- & 96 & 78 & 242 & 65 & 53.7 & 71.3 & 58.4 & 35.5 & 57.9 & 70.5 \\
\bottomrule
\end{tabular}%
}
\caption{Main results. Real-time detectors and our DEYO utilize a consistent input size of 640, while end-to-end detectors employ an input size of (800, 1333). The end-to-end speed results are reported on a T4 GPU with TensorRT FP16, following the method proposed in RT-DETR. We do not test the speed of DETRs, as they are not real time detectors.}
\label{table:1}
\end{center}
\vspace{-4mm}
\end{table*}

\subsection{Setups}

\noindent {\bf COCO} To evaluate the performance of our method in object detection tasks, we conducted experiments on the widely used Microsoft COCO \cite{12}. We trained the DEYO using the \texttt{train2017} and evaluated the performance using the \texttt{val2017}.

\noindent {\bf CrowdHuman} To evaluate the end-to-end effectiveness of DEYO in dense detection compared to classical detectors, we conducted experiments on CrowdHuman \cite{63}. We leveraged the comprehensive full-body annotations available in the dataset and conducted our evaluation on the validation set. In terms of optimizer-related parameters, we adopted the same settings as the COCO. All experimental post-processing is referred to the paper of Iter-Deformable-DETR \cite{69} without any modification.

\begin{table}[t]
\vspace{1.5mm}
\begin{center}
\resizebox{0.47\textwidth}{!}{
\begin{tabular}{lcccccccc}
\toprule
\bf{Method} &\bf{Epochs} &\bf{AP$_5$$_0$} &\bf{mMR} &\bf{Recall} \\
\midrule
ATSS \cite{65} &36 &89.6 &44.4 &95.9\\
DW \cite{66} &36 &89.0 &57.6 &97.4\\
Cascade R-CNN \cite{68} &36 &86.0 &44.1 &89.2\\
Sparse R-CNN \cite{47} &50 &89.2 &48.3 &95.9\\
Deform DETR \cite{7} &50 &89.1 &50.0 &95.3\\
DeFCN \cite{67} &36 &91.0 &46.5 &97.9\\
DEYO-X &300 &\bf{92.3} &\bf{43.3} &97.3\\ \bottomrule
\end{tabular}}
\end{center}
\vspace{-3mm}
\caption{Performance on CrowdHuman (full body).}
\label{table:2}
\vspace{-5mm}
\end{table}

\noindent {\bf Implementation Details} In the first stage of training, we follow \cite{62} the strategy and hyperparameters of training from scratch. In the second stage of training, we used a 6-layer Transformer decoder as the decoder of DEYO. We trained the detector following the \cite{62} hyperparameters, but we used the AdamW \cite{33} optimizer. The learning rate is set to 0.0001, and the weight decay is set to 0.0001. The data enhancement strategy in the second stage is the same as the first stage of training, including random color distortion, inverse translation, flipping, resizing, mosaic, and other operations. On the COCO \cite{12} dataset, except DEYO-tiny, which uses 100 queries, DEYO of other scales uses 300 queries. All evaluations were conducted using a Tesla T4 GPU, complemented by an 8 vCPU Intel Xeon Processor (Skylake, IBRS). The experiments utilized PyTorch version 1.9.0, integrated with TensorRT 8.6.1.

\begin{table*}[t]
\centering
\resizebox{1\textwidth}{!}{%
\setlength{\tabcolsep}{10pt} 
\begin{tabular}{lcccccccc}  
\toprule
\bf{Model} & \bf{Epochs} & \bf{Queries} & \bf{\#Params (M)} & \bf{GFLOPs} & \bf{FPS$_{\text{bs}=1}$} & \bf{AP$_{50}$} & \bf{mMR} & \bf{Recall} \\
\midrule
\rowcolor{gray!20}
\multicolumn{9}{l}{\textit{Classic Object Detectors}} \\
YOLOv8-N \cite{62} & 300 &-- & 9 & 163 &-- & 82.7 & 50.4 & 87.2 \\
YOLOv8-S \cite{62} & 300 &-- & 29 & 143 &-- & 85.4 & 46.0 & 88.2 \\
YOLOv8-M \cite{62} & 300 &-- & 79 & 106 &-- & 86.8 & 43.8 & 89.0 \\
YOLOv8-L \cite{62} & 300 &-- & 44 & 165 &-- & 87.6 & 43.1 & 89.6 \\
YOLOv8-X \cite{62} & 300 &-- & 68 & 258 &-- & 88.1 & 42.9 & 90.0 \\
\midrule
\rowcolor{gray!20}
\multicolumn{9}{l}{\textit{Query-based Object Detectors}} \\
DEYO-N  & 300 &300 & 6 & 10 & 391 & 86.6 & 50.4 & 94.1 \\
DEYO-S  & 300 &300 & 14 & 26 & 296 & 89.3 & 46.6 & 95.2 \\
DEYO-M  & 300 &300 & 33 & 78 & 138 & 91.0 & 44.4 & 96.1 \\
DEYO-L  & 300 &500 & 51 & 158 & 91 & 92.0 & 44.1 & 97.1 \\
DEYO-X  & 300 &500 & 78 & 246 & 62 & 92.3 & 43.3 & 97.3 \\
\bottomrule
\end{tabular}%
}
\caption{Comparative of YOLOv8 and DEYO Performance on the CrowdHuman (full body). Owing to DEYO's abandonment of reliance on NMS, a notable enhancement in performance has been achieved.}
\label{table:3}
\end{table*}
\subsection{Main Results}
\hspace{1pc}We compared the scaled DEYO with YOLOv5 \cite{61}, YOLOv8 \cite{62}, and RT-DETR \cite{58} in Table~\ref{table:1}. Compared to YOLOv8, DEYO significantly improves accuracy by 2.4 AP / 0.9 AP/ 0.5 AP at scales N, S, and M while achieving a 143\% / 110\% / 32\% increase in FPS. At scales L and X, DEYO continues to exhibit a better trade-off between accuracy and speed. As shown in Table~\ref{table:3}, DEYO performs exceptionally well in dense scenarios with real-time speed. Specifically, DEYO-X has attained an impressive 92.3 AP and 43.3 mMR, with a remarkable performance of 97.3 recall within the CrowdHuman \cite{63}.

\begin{table}[t]
\begin{center}
\resizebox{0.47\textwidth}{!}{%
\setlength{\tabcolsep}{15pt} 
\begin{tabular}{lccc}  
\toprule
\bf{Model} & \bf{Epochs} &\bf{AP} & \bf{AP$_{50}$}  \\
\midrule
DEYO-N &12 &35.9 &50.6\\
DEYO-S &12 &43.6 &61.2\\
DEYO-M &12 &49.4 &66.9\\
DEYO-L &12 &51.7 &68.9\\
DEYO-X &12 &52.9 &70.3\\
\midrule
DEYO-N &24 &37.2 &52.4\\
DEYO-S &24 &44.4 &61.2\\
DEYO-M &24 &49.7 &67.3\\
DEYO-L &24 &52.0 &69.5\\
DEYO-X &24 &53.2 &70.7\\
\bottomrule
\end{tabular}%
}
\caption{Results for DEYO coco val2017 trained with more epochs (12, 24).}
\label{table:4}
\end{center}
\vspace{-6mm}
\end{table}

\subsection{Ablation Study}
\hspace{1pc}Table~\ref{table:5} presents the training outcomes for the YOLO \cite{4,5,6} and DEYO models utilizing three distinct training methodologies on the CrowdHuman \cite{63} dataset: the YOLO approach, the DETR approach, and a step-by-step training strategy. The findings indicate that the YOLOv8-N \cite{62} model can achieve an Average Precision (AP) of 82.6, even when trained from scratch without relying on supplementary datasets, by leveraging the abundant supervisory information provided through a one-to-many training strategy. In contrast, the DEYO-N model, constrained by a one-to-one matching training strategy that offers limited supervisory signals, achieved a performance ceiling of 72.1AP despite undergoing the same number of iterations as its YOLO counterpart. Moreover, when the DEYO-N model's backbone was initialized using YOLOv8-N-CLS, pre-trained from ImageNet \cite{59}, and combined with the DETR training strategy, DEYO-N's performance reached 78.3AP. Notably, implementing the step-by-step training significantly enhanced DEYO-N's performance, with an increase of 4.7AP.

\begin{table}[t]
\begin{center}
\resizebox{0.47\textwidth}{!}{%
\setlength{\tabcolsep}{12pt} 
\begin{tabular}{lccc}  
\toprule
\bf{Model} & \bf{Strategy} & \bf{Epochs} &\bf{AP$_5$$_0$}\\
\midrule
\rowcolor{gray!20}
\multicolumn{4}{l}{\textit{Train from Scratch}} \\
DEYO-N &DETR & 300 & 77.2 \\ 
YOLOv8-N \cite{62} &YOLO & 300 & \bf{82.6} \\
\midrule
DEYO-N &DETR & 72  & 78.3 \\
DEYO-N &Step-by-step  & 72  & \bf{83.0} \\
\bottomrule
\end{tabular}%
}
\caption{Comparing different methods trained on the CrowdHuman dataset, it should be noted that for this experiment, we computed the AP50 metric utilizing the tools provided by YOLOv8.}
\label{table:5}
\end{center}
\vspace{-6mm}
\end{table}

In Table~\ref{table:7}, we examined the significance of the high-quality multi-scale features provided by the Neck component, pre-trained in the first phase within the DEYO model. The model's performance markedly decreased by 18.8 average precision points, only achieving 68.3 average precision, when solely utilizing the pre-trained backbone without the pre-trained Neck for step-by-step training. These findings clearly indicate that the key to the DEYO model's superior performance does not lie in the employment of a more sophisticated backbone pre-trained beyond ImageNet but rather in the first-phase pre-trained Neck, which furnishes the model with high-quality multi-scale features.

\begin{table*}[t]
\begin{center}
\resizebox{1\textwidth}{!}{%
\begin{tabular}{lcccccccc}  
\toprule
\bf{Method} & \bf{\#Epochs} & \bf{w/ Step-by-step Training} & \bf{AP} & \bf{AP$_{50}$} & \bf{AP$_{75}$} & \bf{AP$_S$} & \bf{AP$_M$} & \bf{AP$_L$} \\
\midrule
Baseline & 12 & $\times$ & 23.8 & 36.6 & 25.1 & 11.1 & 25.8 & 33.4 \\
Baseline & 12 & $\checkmark$ & 36.3 & 51.4 & 39.1 & 17.3 & 39.9 & 52.2 \\
\midrule
Group-DETR \cite{44} & 12 & $\times$ & 24.1 & 36.3 & 25.6 & 9.8 & 26.0 & 35.4 \\
Group-DETR \cite{44}& 12 & $\checkmark$ & 36.4 & 50.8 & 39.3 & 17.1 & 39.8 & 52.2\\
\midrule
H-DETR \cite{50} & 12 & $\times$ & 24.3 & 36.7 & 25.8 & 10.1 & 26.3 & 35.5 \\
H-DETR \cite{50} & 12 & $\checkmark$ & 36.4 & 50.9 & 39.4 & 17.2 & 39.9 & 52.2 \\
\midrule
DINO \cite{14}& 12 & $\times$ & 24.6 & 36.7 & 25.8 & 10.1 & 26.3 & 35.5 \\
DINO \cite{14}& 12 & $\checkmark$ & 36.5 & 51.1 & 39.5 & 18.3 & 39.9 & 52.2 \\
\bottomrule
\end{tabular}%
}
\caption{Comparison of different methods under the 12-Epoch training setting for DEYO-N. Compared to our approach, previous methodologies could not effectively address the training challenges posed by insufficient supervisory signals, which also resulted in additional training costs.}
\label{table:6}
\end{center}
\vspace{-4mm}
\end{table*}

In Table~\ref{table:8}, we analyze the enhancement resulting from fundamentally addressing the instability of early bipartite graph matching by freezing the DEYO's backbone and neck during the second stage of training, which positively impacts network performance. Compared to fine-tuning the backbone and neck throughout the second phase, the act of freezing yields a 1.1 AP increase in DEYO's performance. Moreover, as the first phase of DEYO involves pre-training on the COCO \cite{12} dataset for the object detection task, it allows for the implementation of more robust data augmentation strategies during the second stage of training. Consequently, unlike DETRs, adopting Mosaic data augmentation does not result in performance degradation; instead, it contributes to a 0.2 AP improvement.

\begin{table}[t]
\begin{center}
\resizebox{0.47\textwidth}{!}{
\setlength{\tabcolsep}{15pt} 
\begin{tabular}{lccccc}
\toprule
\bf{Model} &\bf{Backbone} &\bf{Neck} &\bf{AP$_5$$_0$}\\
\midrule
DEYO-N &$\checkmark$ &$\times$  &68.3\\
DEYO-N &$\checkmark$ &$\checkmark$ &\bf{87.1}\\
\bottomrule
\end{tabular}}
\end{center}
\caption{Results of the ablation study on step-by-step training. (CrowdHuman)}
\label{table:7}
\end{table}

\begin{table}[t]
\begin{center}
\resizebox{0.47\textwidth}{!}{%
\setlength{\tabcolsep}{15pt} 
\begin{tabular}{lcccc}  
\toprule
\bf{Model} &\bf{w/Frozen} & \bf{w/Mosaic} &\bf{AP}\\
\midrule
DEYO-L &$\times$ &$\times$ &51.6\\
DEYO-L &$\checkmark$ &$\times$ &52.5\\
DEYO-L &$\checkmark$ &$\checkmark$ &\bf{52.7}\\
\bottomrule
\end{tabular}%
}
\caption{Exploring the impact of a frozen operation and mosaic data augmentation.}
\label{table:8}
\end{center}
\vspace{-4mm}
\end{table}

\begin{figure}[t]
  \centering
  \includegraphics[width=0.95\linewidth]{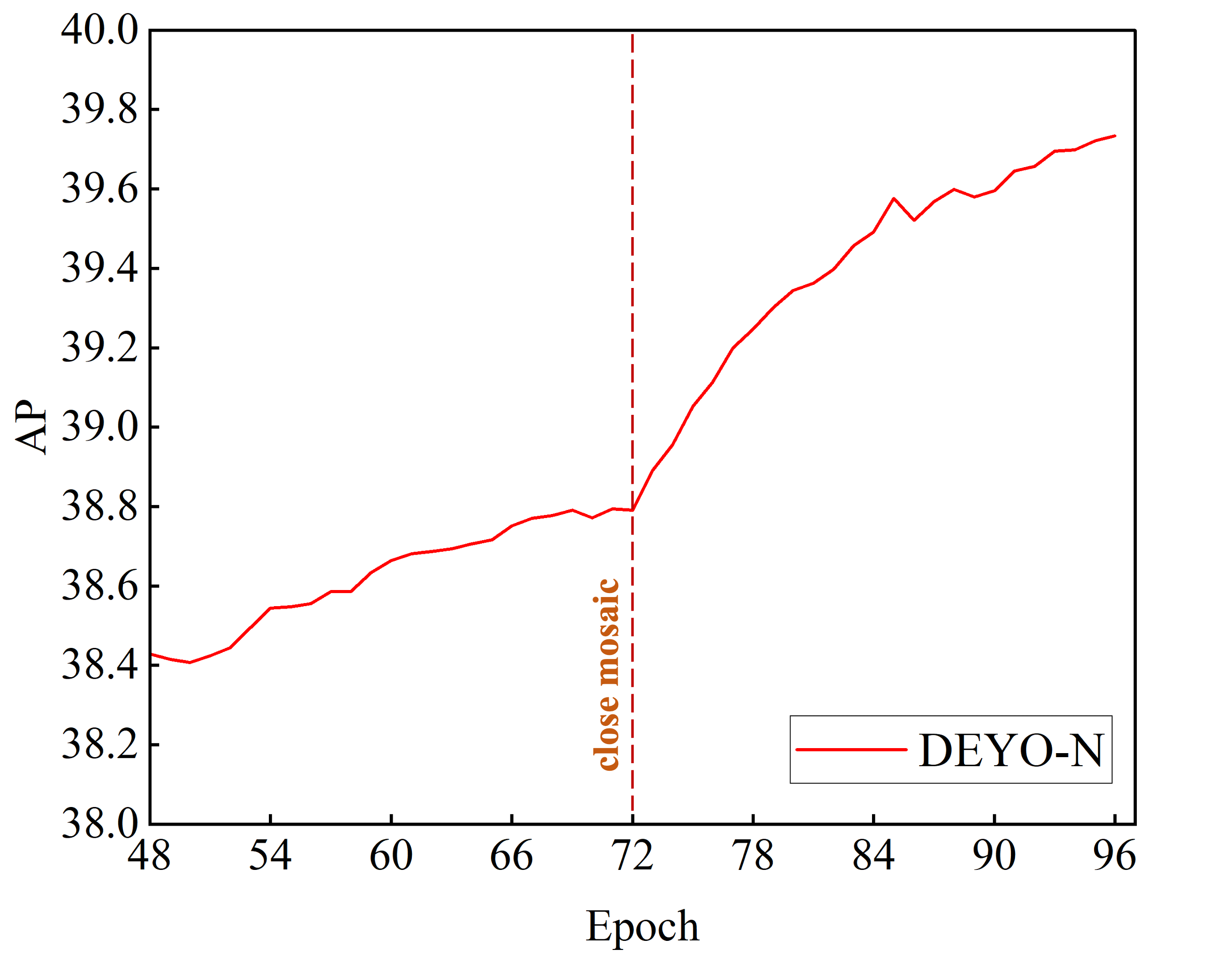}
   \caption{Owing to the initial phase of DEYO being pre-trained on the COCO dataset for the task of object detection, it was afforded the capability to employ more robust data augmentation strategies during the second stage of training. Consequently, DEYO's adoption of the Mosaic data augmentation technique did not result in performance degradation, in contrast to the experience with DETRs.}
   \label{fig:5}
\end{figure}

\begin{figure*}[t]
\begin{center}
\includegraphics[width=0.92\linewidth]{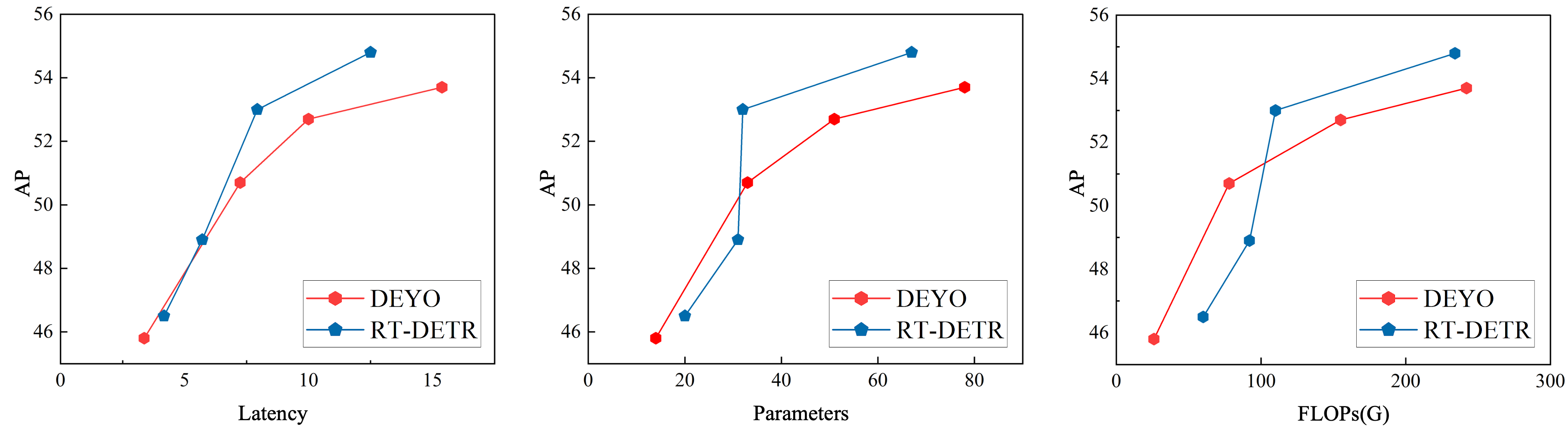}
   \caption{A comprehensive comparison between DEYO and RT-DETR shows that although there is a particular gap on a larger scale, DEYO does not rely on additional training data and has significantly reduced training costs. We believe that DEYO possesses its unique advantages on custom datasets.}
   \label{fig:6}
\vspace{-6mm}
\end{center}
\end{figure*}

\subsection{Analysis}

\hspace{1pc}Experimental outcomes presented in Table~\ref{table:6} elucidate a pronounced degradation in performance for DINO\cite{14}, H-DETR\cite{50}, and Group-DETR\cite{44} when a step-by-step training strategy is not employed, in stark contrast to our method. Compared to our approach, methodologies proposed in references \cite{14, 50, 44} fail to effectively navigate the training challenges precipitated by a lack of supervisory signals while culminating in inflated training expenditures. Conversely, as delineated in Table~\ref{table:9}, our training regimen not only circumvents the imposition of additional training overheads but also significantly curtails the training expenses for the detector (the 3300 queries required by Group-DETR could potentially sextuple the training duration). The initial phase of the DEYO model's training necessitates a mere 16GB of VRAM, while the subsequent phase demands even less, at 8GB of VRAM. For scenarios with constrained training resources, DEYO can deactivate the CDN feature to mitigate VRAM requirements further. As demonstrated in Table~\ref{table:6}, the progressive training strategy affords DEYO a high-caliber pretraining foundation during its first phase, ensuring that even with CDN deactivated, performance detriments remain manageable.

\begin{table}[t]
\vspace{1.5mm}
\begin{center}
\resizebox{0.47\textwidth}{!}{%
\renewcommand{\arraystretch}{1.25}
\begin{tabular}{lccc}  
\toprule
\bf{Model} &\bf{Neck} & \bf{Hidden Dimension} & \bf{GPU Memory} \\
\midrule
YOLOv8-N \cite{62} &(64, 128, 256) &--  &3247MiB\\
YOLOv8-S \cite{62} &(64, 128, 512) &--  &4857MiB\\
YOLOv8-M \cite{62} &(192, 384, 576) &--  &7081MiB\\
YOLOv8-L \cite{62} &(256, 512, 512) &--  &10503MiB\\
YOLOv8-X \cite{62} &(320, 640, 640) &--  &13069MiB\\
\midrule
DEYO-tiny &(64, 128, 256) &64  &2238MiB\\
DEYO-N &(64, 128, 256) &128  &4746MiB\\
DEYO-S &(64, 128, 256) &128 &5062MiB\\
DEYO-M &(192, 384, 576)&256 &6444MiB\\
DEYO-L &(256, 512, 512) &256 &6476MiB\\
DEYO-X &(320, 640, 640)  &320 &6888MiB\\
\midrule
\rowcolor{gray!20}
\multicolumn{4}{l}{\textit{No Contrastive DeNoising Training}} \\
DEYO-tiny &(64, 128, 256) &64 &1514MiB \\
DEYO-N &(64, 128, 256) &128 &2700MiB\\
DEYO-S &(64, 128, 512) &128 &3108MiB\\
DEYO-M &(192, 384, 576) &256 &3948MiB\\
DEYO-L &(256, 512, 512) &256 &4216MiB\\
DEYO-X &(320, 640, 640) &320 &5194MiB\\
\bottomrule
\end{tabular}%
}
\caption{Detailed configurations of YOLO and DEYO as well as their GPU memory usage.}
\label{table:9}
\end{center}
\vspace{-6mm}
\end{table}

As depicted in Fig~\ref{fig:6}, on the X scale, DEYO exhibits some discrepancies when compared to RT-DETR-X, which utilizes pre-training on ImageNet \cite{59}. However, this gap can be attributed to RT-DETR's \cite{58} incorporation of a more efficient backbone. Furthermore, it is our contention that performance on the COCO \cite{12} dataset does not wholly encapsulate the merits and demerits of a detector. Considering that DEYO does not require additional training data, it can leverage more robust data augmentation strategies and incur lower training costs. Consequently, DEYO possesses unique advantages when applied to custom datasets.

Without altering the original backbone and neck of YOLO \cite{4,5,6}, DEYO effortlessly achieved state-of-the-art (SOTA) performance, demonstrating the plug-and-play characteristic of the DEYO model's design philosophy. However, we have observed that the neck of YOLOv8 \cite{62} and the model scaling strategy do not fully align with DEYO. As the model size increases, the performance gains of DEYO diminish incrementally. We postulate that one reason for this is the mismatch between the output dimensions of YOLOv8's neck and the hidden dimensions of DEYO's decoder. This discrepancy underscores the untapped potential of the DEYO model. We believe that a backbone, neck, and model scaling strategy specifically tailored for DEYO and Step-by-step Training could propel DEYO's performance to unprecedented levels.

%% file: sec/5_col.tex
\section{Conclusion}

\hspace{1pc}In this paper, we have innovatively developed a training strategy that not only circumvents the need for additional datasets but also successfully addresses a problem that previous methods could not surmount: performance degradation due to insufficient training of multi-scale feature layers. This approach not only enhances model performance but also significantly reduces training costs. By integrating our meticulously designed lightweight encoder with this revolutionary strategy, we have introduced the DEYO, which surpasses all existing real-time object detectors without relying on supplemental datasets.

We consider DEYO to be a specific instance of the fusion between classic detectors and query-based detectors. We are convinced that other methodologies exist that could satisfy even higher precision requirements. Nevertheless, DEYO's innovative detector design introduces new challenges, such as the need to redesign the backbone and neck to fully realize DEYO's potential. We anticipate that future research will yield effective solutions to these challenges.